\documentclass[times]{elsarticle}

\usepackage{amssymb}
\usepackage{hyperref}
\usepackage{amsmath}
\usepackage{commath}
\usepackage{multirow}
\usepackage{makecell}

\journal{Pattern Recognition}

\begin{document}

\begin{frontmatter}



\title{Robust Detection, Association, and Localization  of Vehicle Lights: A Context-Based Cascaded CNN Approach \& Evaluations}


\author[1]{Akshay Gopalkrishnan\corref{cor1}} 

\cortext[cor1]{Corresponding author:}
\ead{agopalkr@ucsd.edu}

\author[1]{Ross Greer}
\author[1]{Maitrayee Keskar}
\author[1]{Mohan M. Trivedi}

\affiliation[1]{organization={University of California San Diego},
                addressline={Laboratory for Intelligent \& Safe Automobiles}, 
                city={La Jolla}, 
                state={California},
                country={USA}}

\begin{abstract}
Vehicle light detection, association, and localization are required for important downstream safe autonomous driving tasks, such as predicting a vehicle's light state to determine if the vehicle is making a lane change or turning. Currently, many vehicle light detectors use single-stage detectors which predict bounding boxes to identify a vehicle light, in a manner decoupled from vehicle instances. In this paper, we present a method for detecting a vehicle light given an upstream vehicle detection and approximation of a visible light's center. Our method predicts four approximate corners associated with each vehicle light. We experiment with CNN architectures, data augmentation, and contextual preprocessing methods designed to reduce surrounding-vehicle confusion. We achieve an average distance error from the ground truth corner of 4.77 pixels, about 16.33\% of the size of the vehicle light on average. We train and evaluate our model on the LISA Lights Dataset, allowing us to thoroughly evaluate our vehicle light corner detection model on a large variety of vehicle light shapes and lighting conditions. We propose that this model can be integrated into a pipeline with vehicle detection and vehicle light center detection to make a fully-formed vehicle light detection network, valuable to identifying trajectory-informative signals in driving scenes.  
\end{abstract}




\end{frontmatter}


\section{Introduction}
\label{sec:intro}

Detecting car lights is a critical task for autonomous and safe driving, as vehicle lights are key indicators for the future motion of the vehicle. Vehicle lights are constantly used by drivers to indicate to surrounding traffic their future maneuvers or lane changes. As a result, a vehicle light can be used as a cue for models that perform vehicle trajectory predictions \cite{deo2018would} \cite{messaoud2021trajectory} \cite{deo2020trajectory} or combined with driver monitoring data to set up a looking-in and looking-out system \cite{tawari2014looking} to predict driver take-over time\cite{rangesh2021autonomous} \cite{rangesh2021predicting}.  

\begin{figure}
    \centering
    \includegraphics[width=.43\textwidth]{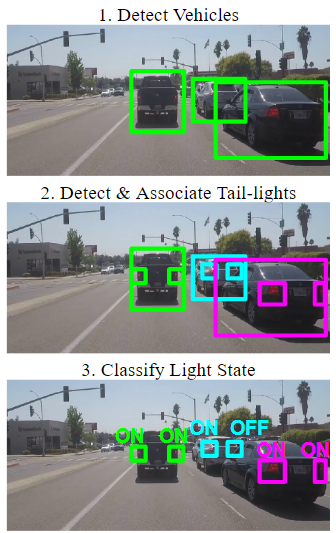}
    \includegraphics[width=.47\textwidth]{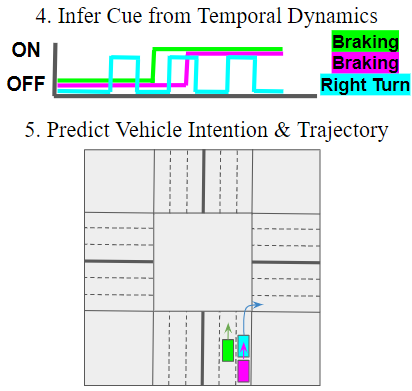}
    \caption{An example multi-stage pipeline for improved trajectory prediction using driver-visible, dynamic cues from the outside scene. In early processing, vehicles and associated tail-lights must be detected, and their states classified. From tracking these states over time, a system can infer the intended cue, which can then inform downstream trajectory prediction models which would otherwise have less information from vehicle position and speed alone. In this research, we present a modular solution for Step 2 in this pipeline.}
    \label{fig:pipeline}
\end{figure}

\begin{figure*}[t]
  \centering
   \includegraphics[width=0.8\linewidth]{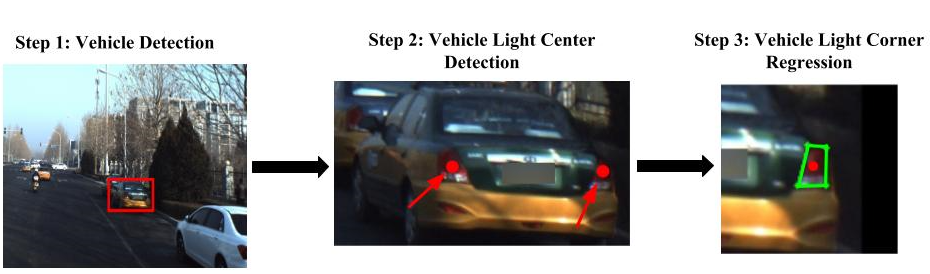}

   \caption{A visualization of the pipeline we propose that uses vehicle light detection and classification as an informative cue for vehicle trajectory predictions.}
   \label{fig1}
\end{figure*}

Accordingly, in this paper, we present a cascaded model approach to vehicle light detection visualized in Figure \ref{fig:pipeline} in which each model performs the following:
\begin{enumerate}
  \item Detect vehicles (2D) in the traffic scene.
  \item From bounding boxes of these detected vehicles, estimate the center of each visible vehicle light.
  \item Using the bounding boxes of vehicles and centers of each visible vehicle light, predict the location of the 4 ``corners" associated with each vehicle light. Note that not all vehicles have a strict ``corner", so this can be taken to refer to a boundary point of two intersecting geometric curves comprising visual edges of the taillight. 
\end{enumerate}
In this research, we focus on developing a model that can solve the third problem defined in this list and also exemplified in the step three of Figure \ref{fig1}. We formulate this problem as a regression task: Given the center coordinate of a car light in an image, we output four $(x,y)$ coordinates which regress from the center of the light to each of the four corners of the vehicle light. We train a CNN on the LISA Lights Dataset \cite{greer2023patterns}, which contains over 40,000 specialized cropped images centered on a vehicle light, to learn features to predict these four $(x,y)$ regression coordinates. Such a model can be cascaded with the preceding model layers to form a complete taillight detector, with the added benefit of implicit association of a detected taillight to its respective vehicle.


\section{Related Research}


\subsection{Approaches to Vehicle light Detection}
 Detecting vehicle lights is especially critical during the night time, as the most visible feature of vehicles in front are the tail and brake lights in such conditions. Malley et al. \cite{o2010vehicle} take advantage of this and use a image processing technique with color thresholding to detect the vehicle tail lights, helping them detect the vehicle itself during night time conditions. Satzoda et al. \cite{satzoda2016looking} similarly aim to detect vehicles at night time using the rear lights as a main cue. To do so, they use an Ada-boost classifier on gray-scale images to detect the vehicle region, segment two sub-regions from the vehicle detection, and then perform red-channel thresholding for taillight detection. Pillai et. al \cite{pillai2016detecting} perform taillight detection for nighttime vehicle detection by using HSV thresholding and connected component analysis to detect and group a taillight pair. Instead of using a thresholding technique, Kosaka \& Ohashi \cite{kosaka2015vision} use an approach coined ``Center Surround Extremas" \cite{agrawal2008censure} for detecting taillights at night. Center Surround Extremas uses an integral image to detect light blobs like taillights moving at high speed.

While the techniques described above achieve high performance for detecting vehicle lights in nighttime conditions,
it is still critical to detect vehicle lights in daytime scenarios as well. Therefore, making vehicle light detectors robust to lighting conditions such as night and day is crucial. Ming \& Kang-Hyun \cite{ming2011vehicle} show that detecting the tail lights can still help for detecting vehicles in the daytime. Their approach uses color segmentation to find horizontal tail light pairs. Cui et al. \cite{cui2015vision} also address this problem of previous approaches, developing a taillight detection framework that works under different illumination circumstances. This framework first detects vehicles with a Deformable Part Model and then extracts the taillight candidates by converting the pixel color space from RGB to HSV to perform color thresholding. Cao et. al \cite{cao2021application} implement a CNN for vehicle detection and then with this cropped vehicle image use the RGB and CMY color spaces for taillight recognition. 

Often, such image processing techniques for day-time conditions are not robust to conditions such as lighting or the distance to the vehicle of interest. For example, in cases where the redness of a taillight is lessened by a shadow, image processing and hand-crafted filters may not provide accurate taillight detections. To address this, deep learning object detection methods can be used to provide robustness to adverse lighting conditions. Rather than just predicting bounding boxes for tail lights, Vancea et. al \cite{vancea2017vehicle} use a FCN based on the VGG16 architecture \cite{simonyan2014very} to perform light segmentation. Since this segmentation network can output more than two taillight clusters, they also add a taillight pair identification step that uses distance and 3-D histogram tests to match red regions from the segmentation output representing the taillights. Vancea et. al mention that when detecting taillights from a far distance, the segmentation produces worse results. Our taillight detection approach is invariant to vehicle distance and in fact uses the vehicle size to constrain the vehicle light prediction sizes. Rampavan \& Ijjina \cite{rampavan2023genetic} focus on brake light detection for motorcycle vehicles using a Mask-RCNN network. They note that two-stage object detectors achieve better performance for smaller objects like brake lights, an important finding in support of similar cascaded approaches to learning component vehicle features. Jeon et. al \cite{jeon2022deep} propose a deep learning cascaded model with a lane detector, car detector, and taillight detector. For specifically the taillight detector module, they use a Recurrent Rolling Convolution architecture to find the taillight regions of a vehicle. 

We note that most papers ignore the detection of the front lights, which are still an important feature to detect in many driving scenarios. For example, consider a four-way intersection with stop signs on each side. Detecting the front lights of the incoming vehicles would be critical to see if they are turning and therefore potentially interfering with the ego-vehicle's trajectory. Our approach addresses this issue and presents a model that can detect both front and tail lights. Moreover, similar to other traffic detection problems that involve non-rectangular objects such as traffic lights and signs \cite{mogelmose2015detection, philipsen2015traffic}, previous research treats vehicle light detection as predicting a bounding box even though almost all vehicle lights are not a perfect rectangle. In this paper, we introduce a more precise vehicle light detection model that can predict and accurately fit a variety of vehicle light shapes. 

As stated by Rapson et al. \cite{rapson2019performance}, the task of detecting car lights is difficult for a variety of reasons including: 
\begin{itemize}
    \item variety of car light shapes and brightness
    \item variety of occlusions and orientations 
    \item environment lighting conditions (one of the most common problems in camera related Automated Driving Systems). 
\end{itemize} 
Deep learning methods have been effective in solving image recognition and detection tasks where the objects of interest may appear with such variation, in particular through the use of CNNs. Many vehicle light detection models use end-to-end object detection approach to predict a bounding box around the head or taillight \cite{li2020highly, vancea2017vehicle, rampavan2023genetic}. However, for certain tasks, data preprocessing and feature extraction can provide stronger performance by relieving the learning algorithm of the challenge of discovering features which human experts already know to be important to the task at hand \cite{dziezyc2020can}. Further, explicitly engineered (versus implicitly learned) features assist towards AI system explainability \cite{gosiewska2021simpler, shevskaya2021explainable}, a growing concern for safety-critical systems such as autonomous vehicles \cite{zablocki2022explainability}.

\subsection{Corner-based 2D Object Detection}
Most 2D object detection modules predict a bounding box around the object of interest. There are a few ways to encode a 2D box with four pieces of information, and the most typical encodings utilize (1) a 2D origin point, a height, and a width, or (2) two 2D points. One benefit of the prediction of keypoints (e.g. Law \& Deng's CornerNet \cite{law2018cornernet}) is that the boundary features of the object may be easier to identify as singular points, rather than learning the span of an object through anchor-box learning in methods like Faster RCNN \cite{girshick2015fast} or YOLO \cite{redmon2016you}. Dörr et al. \cite{dorr2022tetrapacknet} base their approach off of CornerNet with TetraPackNet, but instead of predicting a top-left corner and bottom-right corner an object is represented as four arbitrary vertices. Zhou et al. \cite{zhou2019objects} take an entirely different approach to object detection compared to 2D bounding boxes, instead viewing an object as collection of interconnected keypoints, developing CenterNet, which models an object by a single center point (presumed to be akin to the center of a traditional 2D bounding box). Using this center point, the detector can regress to other properties like the object corners, orientation, or even pose. We can see, that prediction of four corners of a bounding box (e.g. \cite{dorr2022tetrapacknet} and this research, in which we predict the four corners of the light as four $(x,y)$ distances from the known light center) represents a hybrid method, where specific, fixed landmark points of interest are being detected (rather than a broad parameterization of a box as height, width, and origin, for which many possible encodings exist which define the same box). Zhao et al. \cite{zhao2022corner} improve CenterNet by introducing CenternessNet. This detector adds box-edge length constraints to CenterNet which improves its ability to differentiate the corners of objects in the same category and reduces the computational expenses of CenterNet as well. 

CenterNet and its derivatives have further utility upstream in the proposed cascaded model, as identifying the center of the vehicle would allow for a variety of regressions to points of interest (light centers, light corners, and other key features). While we limit the scope of this particular research to the identification of light corners from light centers, we highlight the utility of this center-based detection approach toward similar associated tasks in vehicle detection. 

\section{Methods}
\begin{figure}
  \centering
   \includegraphics[width=1\linewidth]{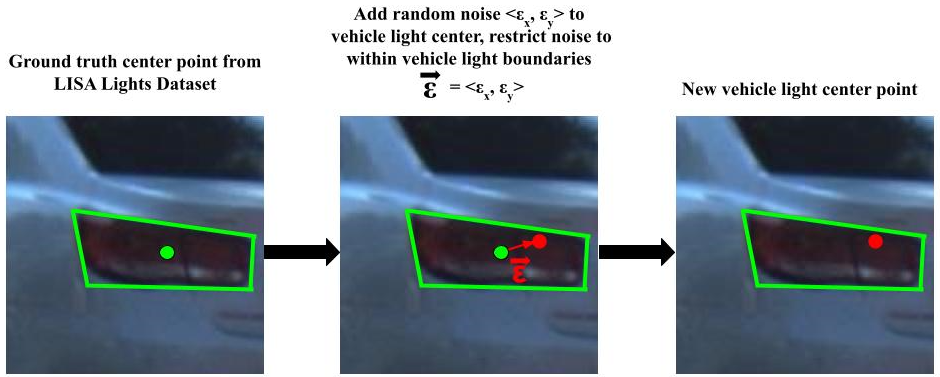}

   \caption{The process for taking a ground-truth vehicle light center point from the LISA Lights Dataset \cite{greer2023patterns} and adding random noise to the center point.}
   \label{fig:noise}
\end{figure}

\begin{figure}
  \centering
   \includegraphics[width=1\linewidth]{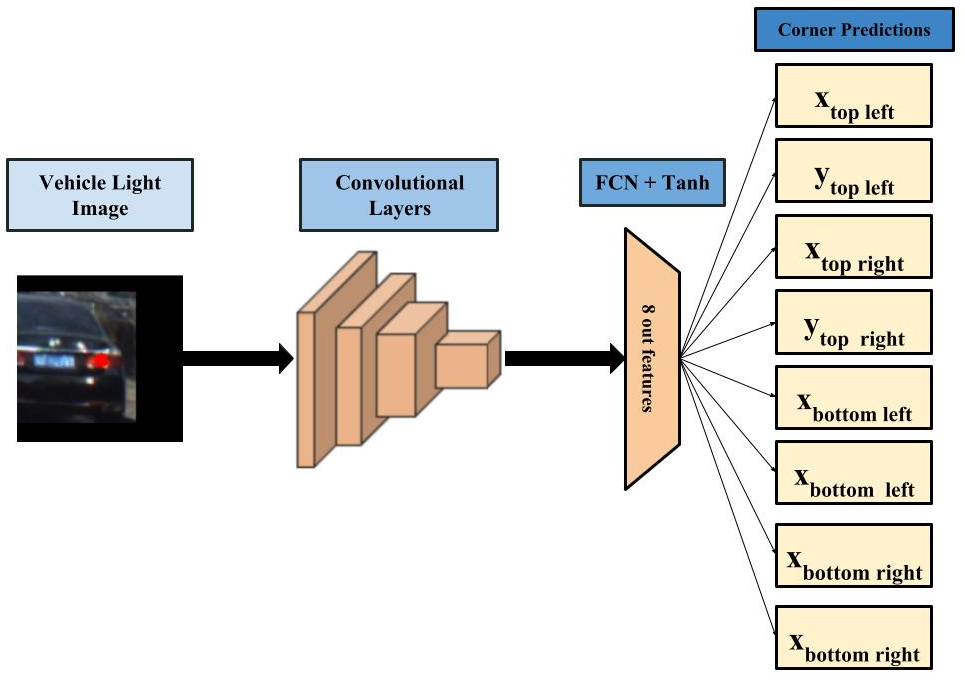}

   \caption{The pipeline for the single vehicle light models to make the four corner predictions. The single vehicle light models take in a vehicle light image from the LISA Lights Dataset as input for a CNN that predicts the (x,y) distance from the vehicle light center to the four corners of the vehicle light.}
   \label{fig3}
\end{figure}

\begin{figure}
  \centering
   \includegraphics[width=1\linewidth]{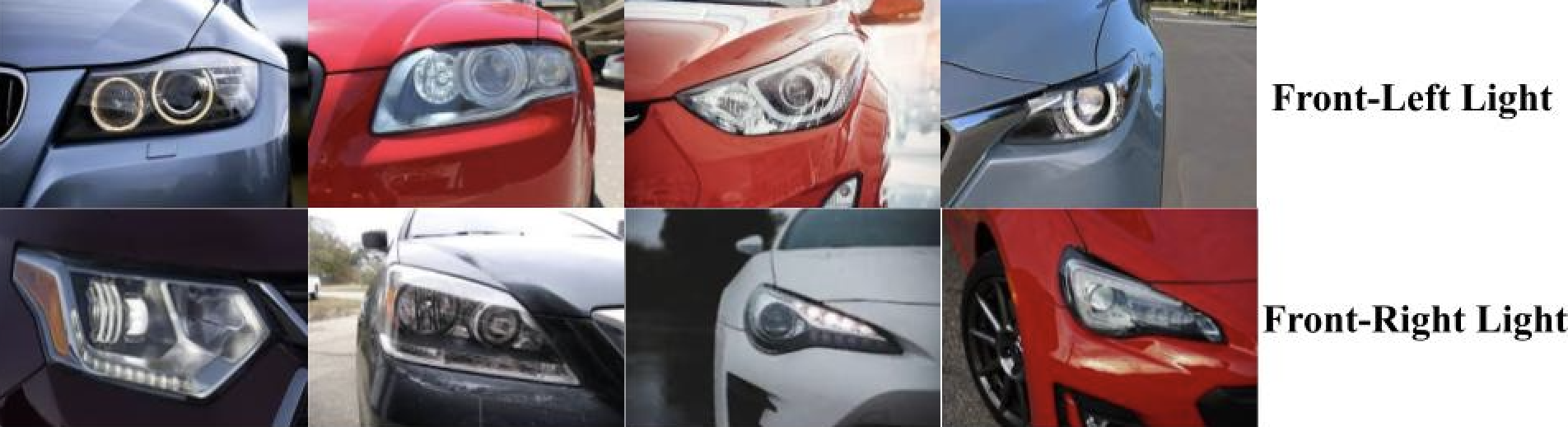}

   \caption{Images of a variety of the left and front lights. Comparing to Figure \ref{rear_lights}, the front lights have different colors and shapes. Furthermore, the front-left and front-right lights are oriented differently, demonstrating the need for a separate model for each of the front lights.}
   \label{front_lights}
\end{figure}

\begin{figure}[hbt!]
  \centering
   \includegraphics[width=1\linewidth]{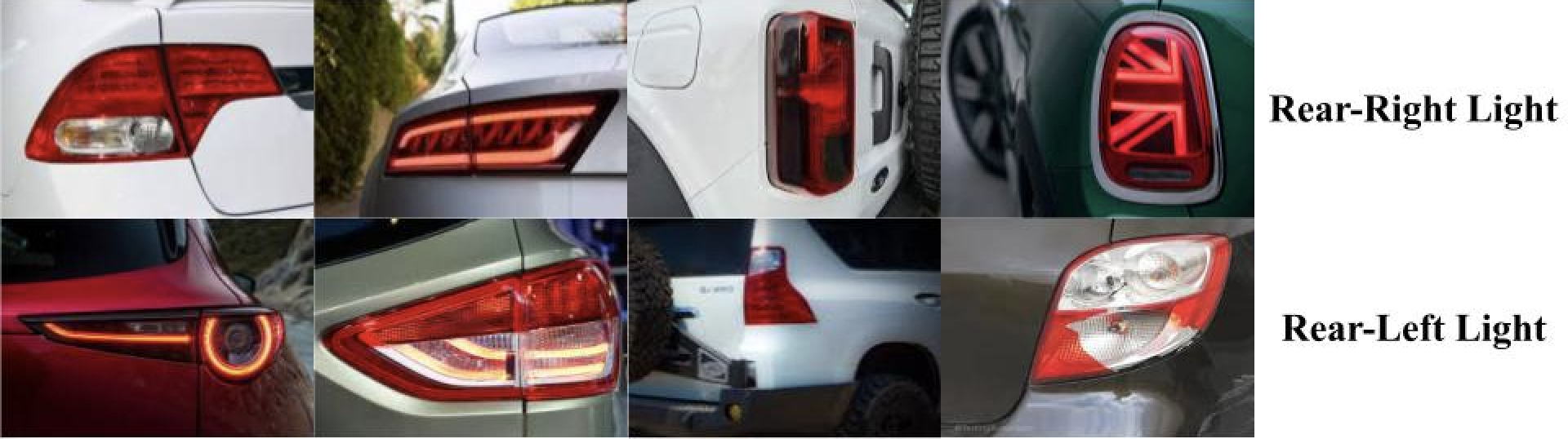}

   \caption{Images of different left and rear vehicle lights. There are a variety of shapes for the rear lights that differ from the front lights shown in Figure \ref{front_lights}. Consequently, a single model designed to make vehicle light corner predictions on all four types of lights would struggle to make accurate predictions.}
   \label{rear_lights}
\end{figure}

 To train a model to predict the corners of a vehicle light, we use the LISA Lights Dataset \cite{greer2023patterns}. This dataset contains over 40,000 images of the four different vehicle lights with special images cropped around the ground-truth center of the vehicle light. In addition, the (x,y) locations of each visible corner for a vehicle light are provided in the dataset, allowing us to train a CNN that can extract features from these cropped vehicle light images to predict corner points. In the cascaded model approach we propose to detect vehicle lights, the CNN models designed in this paper rely on a center estimation of each visible vehicle light. Therefore, this model must be robust to vehicle light center prediction noise and not solely depend on the vehicle light center ground truths provided in the LISA Lights Dataset. To attempt to address this, we add random offset noise to the vehicle light centers to a subset of the LISA Lights Dataset and train separate CNN models on this new dataset with vehicle light center point noise. Given a ground-truth vehicle light center with coordinates $(x_{\text{center}}, y_{\text{center}})$, we add some random noise $(\epsilon_x, \epsilon_y)$ to the ground-truth vehicle light center, giving us a new vehicle light center coordinate of $(x_{\text{center}} + \epsilon_x, y_{\text{center}} + \epsilon_y, )$. We then center the vehicle light image around this noisy center point and used this cropped image around the vehicle light as the image we use to train the CNN models. The distribution used to generate this random noise reflects actual errors that are made when predicting a vehicle light center. Therefore, in some examples there may be zero noise added, as there are cases when the vehicle light center is accurately predicted. Figure \ref{fig:noise} highlights this process to add random noise to the vehicle light centers.

 We modify standard CNN architectures (Resnet \cite{he2016deep}, Densenet \cite{huang2017densely}) by reducing the output size of the final layer to 8 values and using the tanh activation function to normalize predictions to a scale of -1 to 1. 
We test our network on pretrained versions of Resnet-18, Resnet-34, Resnet-50, Resnet-101, DenseNet-121, and DenseNet-169 to evaluate which model performs best for this task. We begin each model from pretrained weights optimized to the ImageNet classification task \cite{deng2009imagenet}. The corner regression model is then trained using a custom MSE loss that we define below:

\begin{equation}
    \mathcal{L} = \frac{1}{N}\sum_{i=1}^N\frac{1}{V_i}\sum_{j=1}^{4} \norm{p_{ij}M_{ij}-t_{ij}}_2
    \label{eq1}
\end{equation}

In this equation, N represents the number of examples, $V_i$ represents the number of visible corners for the \emph{i}th example, and $p_{ij}$ and $t_{ij}$ are the (x,y) regression predictions and targets for the \emph{j}th corner of the \emph{i}th example. $M_{ij}$ is a boolean mask which is $10^{-8}$ if $t_{ij}=0$ (corner is not visible) and 1 otherwise. We avoid setting $M_{ij}=0$ if $t_{ij}=0$ because this would cause the weights to infinitely increase their value during backpropagation. This alteration of regression loss avoids penalizing the corner regression predictions for non-visible corners. While we originally used one CNN regression model to predict the corners for all vehicle lights, we found that training four separate models to learn corner prediction for the front left light, front right light, rear left light, and rear right light improved performance. As shown by Figure \ref{front_lights} and \ref{rear_lights}, the rear and front lights have significantly different shapes and colors, so it would be more difficult for a single model to generalize corner predictions for these type different vehicle lights. Instead, a model trained on a singular type of vehicle light type will be able to learn the features solely pertaining to this light and make more accurate corner predictions. Figure \ref{fig3} highlights the process defined above to make these corner predictions for a single vehicle light model. 

We use the Adam optimizer \cite{kingma2014adam} to update our weights, Stochastic Weighting Averaging with a learning rate decay, a learning rate of $10^{-3}$, weight decay of $10^{-4}$, and train for 25 epochs for each vehicle light model. 

\section{Experimental Analysis and Evaluations}

\subsection{Quantitative Analysis}

We evaluate the performance of our separate vehicle light models based on three metrics: the custom regression loss defined in the previous section, the average distance error (ADE) between a regression prediction and target, and the average percent error of a corner prediction. To calculate the average distance error we use the following the equation: 
\begin{equation}
    ADE = \frac{1}{4}\sum_{i=1}^4\frac{1}{V_i}\sum_{j=1}^{N} 64\times\norm{p_{ij}M_{ij}-t_{ij}}_2
\end{equation}

The Average Distance Error uses the same variables defined in Equation \ref{eq1}. We multiply the distance between the predictions and targets by 64 since both are normalized from -1 to 1 with respect to 64, so multiplying by 64 will give us the actual pixel distance between the prediction and target in the image. This metric represents the average pixel distance for every corner label in our dataset. The average percent error metric is calculated as following:
\begin{equation}
    \text{\% Error} = \frac{1}{N}\sum_{i=1}^N\frac{1}{V_i}\sum_{j=1}^{4} \frac{\norm{p_{ij}M_{ij}-t_{ij}}_2}{\sqrt{W_{ij}^2 + H_{ij}^2}}
\end{equation}
We again use similar variables to Equation \ref{eq1} but introduce $W_{ij}$ and $H_{ij}$, which represent the width and height for the \emph{jth} corner of the \emph{ith} example respectively. $\sqrt{W_{ij}^2 + H_{ij}^2}$ represents the maximum distance a corner prediction can be while staying inside of the vehicle light box. Therefore, a percent error greater than 100\% means a prediction is out of range of the taillight corners. Some autonomous driving detection tasks such as detecting signs emphasize detection on important objects relative to the driver \cite{greer2022salience, greer2023salient, ohn2017all}. We use a similar approach to determine the overall performance of our models through a weighted average of the separate vehicle model performance metrics based on each vehicle light test set size. We also compare the performance of models that are trained using the LISA Light Dataset \cite{greer2023patterns} ground-truth vehicle light centers and the models that are trained with the added random noise to the ground-truth vehicle light centers. 

\begin{figure}
    \centering
    \includegraphics[width=\textwidth]{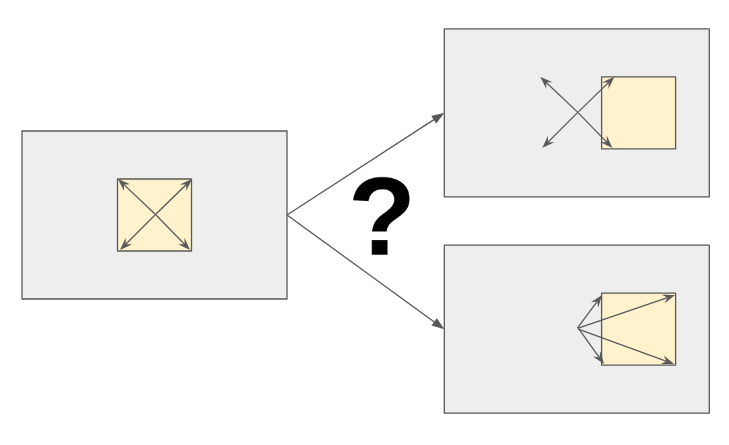}
    \caption{When trained on images like the one on left, in noisy conditions shown on right, which prediction will the model make? In other words: how practical is such a detection and localization model in the real world? A key consideration of this cascaded approach is that it first relies on detection of vehicle light centers, which will have its own margin of error. While in the ideal case (as shown on left and used during our training), the light (yellow) will be center-aligned within the input image (gray), this will not always be the case. With a model trained on such ``ideal" samples, our experiments are designed to assess whether the model has effectively learned to decouple the concept of ``center" from the features learned from the image (bottom right), learning to predict the correct regions of interest from the image whether or not that image is aligned to the center location it was trained on. Given these considerations, we evaluate on two test sets: one with ``ideally"-centered lights and one with random (but naturalistic) offsets of the vehicle light within the image, assessing if the model learns robust to this type of noise.}
    \label{fig:question}
\end{figure}

\begin{table}
\centering
\caption{Comparative performance of various CNN architectures on the ground-truth vehicle light center points test set.}
\begin{tabular}{|c|c|c|c|c|}
\hline
\textbf{Model Type}  & \textbf{Regression Loss} & \textbf{ADE} & \textbf{\% Error}\\
\hline \hline
\multicolumn{4}{|c|}{\textbf{Trained with ground-truth vehicle light center points}} \\
\hline
ResNet-18 &  0.091 & 5.68 & 19.97 \\
\hline
ResNet-34 & 0.081 & 4.99 & 17.22 \\
\hline
ResNet-50 & 0.080 & 4.94 & 16.58 \\
\hline
\textbf{ResNet-101} & \textbf{0.077} & \textbf{4.77} & \textbf{16.33} \\
\hline
DenseNet-121 & 0.081 & 4.98 & 16.68\\
\hline
DenseNet-169 & 0.093 & 5.78 & 20.22 \\
\hline
Vision Transformer (ViT) & 0.081 & 5.18 &  17.28\\
\hline
\multicolumn{4}{|c|}{\textbf{Trained with added noise to ground-truth vehicle light center points}} \\
\hline
ResNet-50 & 0.086 & 5.31 & 17.88 \\
\hline
ResNet-101 & 0.088 & 4.58 & 18.46 \\
\hline
\end{tabular}
\label{CNNStructures}
\end{table}

Table \ref{CNNStructures} highlights the performance on the LISA Lights test set of various CNN architectures trained on either the ground-truth vehicle light center points of the LISA Lights Dataset or with added noise to the vehicle light center points. ResNet-101 achieves the best performance in all three metric categories. As we increase model complexity up until ResNet-101, the performance improves. However, the model performance starts to decline as we use even more complex models such as DenseNet-121. This suggests that demonstrates that Resnet-50 or Resnet-101 offers the best model complexity for our vehicle light corner regression task. In addition, we are also interested to analyze how current state-of-the art models such as a Vision Transformer \cite{dosovitskiy2020image} would perform for this task. From these results, we can see that the Vision Transformer (ViT) leads to an approximate 1\% error increase in comparison to the best performing models such as ResNet-50 and ResNet-101. The models trained on data with added random noise to the ground-truth vehicle light center points performed worse than the models trained on the ground-truth vehicle light center points from the LISA Lights Dataset. With the added noise to the training set, these models may not be as accustomed to vehicle light images without center-point noise, which may lead to their worse performance.

\begin{table}
\centering
\caption{Comparative performance of various CNN architectures on the vehicle light corner regression task given ground-truth vehicle light center point test set with added noise.}
\begin{tabular}{|c|c|c|c|c|}
\hline
\textbf{Model Type}  & \textbf{Regression Loss} & \textbf{ADE} & \textbf{\% Error}\\
\hline \hline
\multicolumn{4}{|c|}{\textbf{Trained with ground-truth vehicle light center points}} \\
\hline
ResNet-18 &  0.080 & 5.12 & 20.54 \\
\hline
ResNet-34 & 0.076 & 4.84 & 16.76 \\
\hline
ResNet-50 & 0.071 & 4.52 & 16.18 \\
\hline
ResNet-101 & \textbf{0.070} & \textbf{4.51} & 16.18 \\
\hline
DenseNet-121 & 0.071 & 4.56 & \textbf{16.16} \\
\hline
DenseNet-169 & 0.078 & 4.98 & 18.07 \\
\hline
Vision Transformer (ViT) & 0.081 & 5.15 & 17.32 \\
\hline
\multicolumn{4}{|c|}{\textbf{Trained with added noise to ground-truth vehicle light center points}} \\
\hline
ResNet-50 & 0.075 & 4.80 & 16.68 \\
\hline
ResNet-101 & 0.074 & 4.72 & 16.46 \\
\hline
\end{tabular}
\label{CNNStructures_noise}
\end{table}

In addition, we also record the performance of various CNN architectures on the LISA Lights test set with added noise, which is presented in Table \ref{CNNStructures_noise}. While we hypothesized that the ResNet-50 and ResNet-101 models trained with added noise would outperform the ground-truth trained models on a test set with added center point noise, the majority of the models trained with ground-truth vehicle light center points achieve a higher performance. To train an effective model with added noise to the ground-truth vehicle light centers, more complex models that can learn both the correct location of the vehicle light center point and the vehicle light corners may be required. In addition, a larger dataset with a variety of vehicle light center point noise will also make a more robust dataset to perform training and evaluation. We also note that models trained on ground-truth vehicle light center points in general achieve higher performance on the test set with added noise. This may demonstrate that consistently providing an accurate ground-truth vehicle light center point in training strengthens the model's understanding of the location of a vehicle light center point during inference time.

\begin{table}[hbt!]
\centering
\caption{Regression performance for different vehicle light cropping approaches.}
\begin{tabular}{|c|c|c|c|}
\hline
\textbf{Cropping Vehicle Light Approach } & \textbf{Regression Loss} & \textbf{ADE} & \textbf{\% Error} \\
\hline \hline
Vehicle with Scene Context Approach & 0.085 & 5.26 & 18.48 \\
\hline
\textbf{Vehicle-Only Context Approach} & \textbf{0.077} & \textbf{4.77} & \textbf{16.33} \\
\hline
\end{tabular}
\label{table:vehcontext}
\end{table}
We also evaluate the performance of the two different vehicle light image generation processes the LISA Lights Dataset provides us: the Vehicle with Scene Context Approach and the Vehicle-Only Context Approach. As explained in the LISA Lights Dataset paper \cite{greer2023patterns}, the Vehicle with Scene Context Approach crops the vehicle light using the full traffic scene image while the Vehicle-Only Context Approach crops the vehicle light from a cropped image of the vehicle. To place the center of the vehicle light in the middle of the image, the Vehicle-Only Context Approach may also add black padding. For this experiment, we use ResNet-101 for each of our vehicle light corner regression models and train with the ground-truth vehicle light center points. We can see that the approach which uses a degree of meaningful feature extraction (i.e. using only the vehicle in the crop) achieves a better regression loss and average distance error than a more ``end-to-end" approach cropping from the full scene visual context. Knowing the image of the vehicle helps detect the vehicle light since using the vehicle image filters out noise from the traffic scene and surrounding vehicles. In addition, the black padding, which eliminates any unnecessary traffic scene information, can constrain the regression predictions of the model so it can learn to never make predictions that will make the taillight corner locations outside the vehicle region; the vehicle-context-only model contains a more strongly constrained foreground. This simplifies the learning task for the CNN regression model further and allows for more accurate vehicle light predictions. 

\begin{table}[hbt!]
\caption{Regression performance for models trained on the regular vehicle light dataset and the expanded vehicle light dataset with augmentations (horizontal reflections).}
\centering
\begin{tabular}{|c|c|c|c|}
\hline
\textbf{Dataset Type} & \textbf{Regression Loss} & \textbf{ADE} & \textbf{\% Error}\\
\hline \hline
\textbf{LISA Lights Dataset} & \textbf{0.077} & \textbf{4.77} & \textbf{16.33}  \\
\hline
LISA Lights Dataset with Augmentations & 0.080 & 4.88 & 16.61 \\
\hline
\end{tabular}
\label{table:augmentation}
\end{table}
To analyze if data augmentation through horizontal reflections improves the model accuracy, we use a Resnet-101 and trained it on the regular LISA Lights dataset and the LISA Lights dataset with augmentations (horizontal reflections of vehicle light images). Both datasets use the ``Vehicle-Only Context Approach`` as this was shown to improve the vehicle light corner regression performance. As shown in Table \ref{table:augmentation}, the expanded LISA Lights Dataset with a horizontal reflection of each vehicle light achieves similar performance to the regular LISA Lights Dataset. While it is true that lights have a chirality (the left-hand light cannot be superimposed onto the right-hand light), the visual features that distinguish a light from a non-light background would be expected to be similar regardless of this reflection. However, it is important to note that reflecting does strongly influence the imagined ``perspective" from which the camera views the vehicle. So, it could be the case that these reflections create artificial viewing angles which are not common to driving patterns where such data is collected, essentially creating virtual car orientations that the machine has no practical use in learning (leading to unimproved performance on the real-world-only test dataset). 
\begin{table}[hbt!]
\centering
\caption{Regression performance for each of the single vehicle light models.}
\begin{tabular}{|c|c|c|c|}
\hline
\textbf{Vehicle Light} & \textbf{Regression Loss} & \textbf{ADE} & \textbf{\% Error} \\
\hline \hline
Left-Front & 0.079 & 4.89 & \textbf{14.15} \\
\hline
Left-Rear & \textbf{0.072} & \textbf{4.42} & 15.71 \\
\hline
Right-Front & 0.91 & 5.48 & 20.46 \\
\hline
Right-Rear &  0.080 & 4.91 & 17.29\\
\hline
\end{tabular}
\label{table:lights}
\end{table}

Table \ref{table:lights} presents the performance of each vehicle light model. For all statistics, the left-front and left-rear corner prediction models perform the best. There is a significant amount of data for these lights, which provided the light models enough examples to make accurate corner predictions. We can attest the lower performance of the front-right light to the smaller dataset we have for this light (4,452 images). Collecting more examples for the front-right light so there is a similar amount of images as the other lights would significantly improve performance and model generalization.


\begin{table*}[hbt!]
\centering 
\caption{Comparison of vehicle light detection performance of our vehicle light detection method to other taillight detection approaches. We note that some methods describe their performance on videos, in which cases the number of frames of the video is unspecified in the cited research.}
\resizebox{\textwidth}{!}{%
\begin{tabular}{|c|c|c|c|c|}
\hline
\thead{\textbf{Vehicle Light} \\ \textbf{Detection Approach}} & \thead{\textbf{Evaluation}\\ \textbf{Dataset}} & \thead{\textbf{\# of Evaluated}\\ \textbf{Images}} & \thead{\textbf{Performance}\\ \textbf{Metric/s}} & \thead{\textbf{Performance} \\ \textbf{Metric Results}} \\
\hline \hline
Rapson et al. \cite{rapson2019performance} & \href{https://cerv.aut.ac.nz/vehicle-lights-dataset/}{\footnotesize{Vehicle Lights Dataset}} & 1,869 & mAP@25, mAP@50 & 18, 5\\
\hline
Vancea et al. \cite{vancea2017vehicle} & \footnotesize{KITTI Tracking (subset)} & 3 videos & Segmentation Accuracy & 95.8 \\
\hline
Jeon et al. \cite{jeon2022deep} & \footnotesize{KITTI Tracking (subset)} & 3 videos & mAP@50 & 100.0 \\
\hline
\textbf{Our Approach} & \href{https://cvrr.ucsd.edu/vehicle-lights-dataset}{\footnotesize{LISA Lights Dataset}} &  \textbf{2,464} & mAP@25, mAP@50 & 97.87, 84.15 \\
\hline
\end{tabular}
}
\label{table:iou}
\end{table*}

Furthermore, we compare the performance of our vehicle light detection approach to other taillight detection methods in research. We apply the same metrics Rapson et al. use, mAP@25 and mAP@50, to evaluate the performance of our model on the dataset we use. To do so, we treat our corner predictions and ground truths as a bounding box and calculate the IOU between the prediction and ground truth box. If the IOU was greater than a threshold $\alpha$, which is 0.25 for mAP@25 and 0.5 for mAP@50, then we would count this prediction as a correct prediction. We note that Vancea et al. \cite{vancea2017vehicle} and Jeon et al. \cite{jeon2022deep} report these metrics while evaluating on three different videos from the KITTI tracking dataset. Our approach achieves similar results to current taillight detection methods and also includes a more thorough evaluation of our model given the amount of traffic scene images we evaluated on. In addition, we detect both front and rear lights at a high accuracy, while these other methods besides Rapson et al. solely focus on taillight detection. While we are not performing a full-scale vehicle light detection process, these results demonstrate that our model can be integrated with a vehicle detector and vehicle light center detector model and still achieve strong performance in these metrics.

\subsection{Qualitative Analysis}

\begin{figure*}[hbt!]
    \centering
    \includegraphics[width=.2\textwidth]{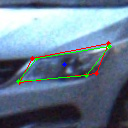}
    \includegraphics[width=.2\textwidth]{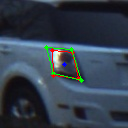}
     \includegraphics[width=.2\textwidth]{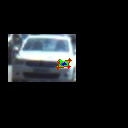}
    \includegraphics[width=.2\textwidth]{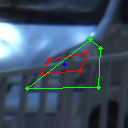}
    \includegraphics[width=.2\textwidth]{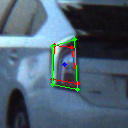}
    \includegraphics[width=.2\textwidth]{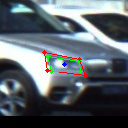}
     \includegraphics[width=.2\textwidth]{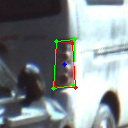}
    \includegraphics[width=.2\textwidth]{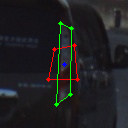}
     \includegraphics[width=.2\textwidth]{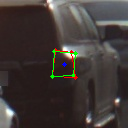}
      \includegraphics[width=.2\textwidth]{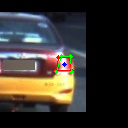}
       \includegraphics[width=.2\textwidth]{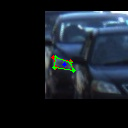}
        \includegraphics[width=.2\textwidth]{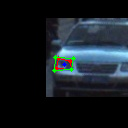}

    \caption{Qualitative results of the vehicle light corner regression predictions from the original test set, with vehicle light center points aligned to image center. The blue dot represents the center of the vehicle light, the green outline represents the ground truth shape of the vehicle light, and the red outline represents the predicted shape of the vehicle light.}
    \label{fig:qual}
\end{figure*}

In addition to the performance metrics, we also visualize our model's predictions on examples from the LISA Lights Dataset. To generate the predictions, we use a ResNet-101 trained on the ground-truth vehicle light center points of the LISA Lights dataset. Figure \ref{fig:qual} shows some arbitrarily selected predictions from the ground-truth vehicle light center test set of our model, meant to be representative of the spectrum of performance. Overall, it predicts shapes similar to the ground truth and is robust to different types of vehicle lights. Furthermore, the light detector is still effective in a variety of adverse conditions such as irregular lighting and far distances from the vehicle of interest. However, for some irregular tail light shapes or occluded vehicle lights (such as the images in the 1st and 2nd row of the 3rd column of Figure \ref{fig:qual}), our model does not predicts a shape similar to the ground truth. This is expected as some of these cases of lights are irregular and offer less examples for the model to learn from. 

\begin{figure*}[hbt!]
    \centering
    \includegraphics[width=.2\textwidth]{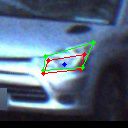}
    \includegraphics[width=.2\textwidth]{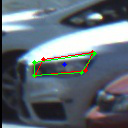}
     \includegraphics[width=.2\textwidth]{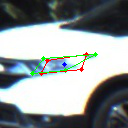}
    \includegraphics[width=.2\textwidth]{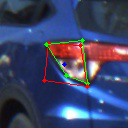}
    \includegraphics[width=.2\textwidth]{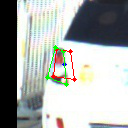}
    \includegraphics[width=.2\textwidth]{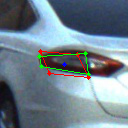}
     \includegraphics[width=.2\textwidth]{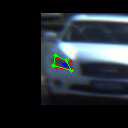}
    \includegraphics[width=.2\textwidth]{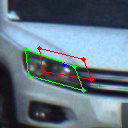}
    \includegraphics[width=.2\textwidth]{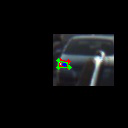}
    \includegraphics[width=.2\textwidth]{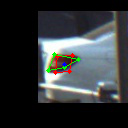}
    \includegraphics[width=.2\textwidth]{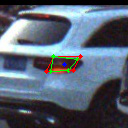}
    \includegraphics[width=.2\textwidth]{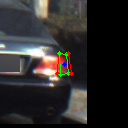}

    \caption{Qualitative results of the vehicle light corner regression predictions from the test set with added center-point noise. The blue dot represents the center of the vehicle light, the green outline represents the ground truth shape of the vehicle light, and the red outline represents the predicted shape of the vehicle light.}
    \label{fig:noise_qual}
\end{figure*}

Figure \ref{fig:noise_qual} provides example predictions from the vehicle light model on the test set with added center-point noise. Similar to Figure \ref{fig:qual}, these predictions were chosen to be representative of the model performance. We again use ResNet-101 trained on the ground-truth vehicle light center points, as this was the best performing model on the test set with added center-point noise. Even with the center point noise added, the model still makes accurate predictions that encapsulate the majority of the vehicle light. There are cases such as the fourth image in the second row of Figure \ref{fig:noise_qual} where the model is unable to adjust to a larger amount of vehicle light center-point noise. The quantitative performance discussed earlier as well as these qualitative results highlight how the ground-truth vehicle light center points can still be used to train a model robust to vehicle light center-point noise.

\section{Concluding Remarks}
In this paper, we have presented a corner-based method of 2D vehicle light detection using a CNN model that takes as input a given vehicle detection and visible light center point, and predicts as output the corners of the vehicle light. Importantly, this method implicitly learns to associate detected lights with a particular vehicle due to the cascaded model approach which integrates prior information of vehicle detections (while also voiding non-vehicle context from the scene image).

Why are these vehicle lights important and relevant to a nearby autonomous vehicle? Beyond the role of vehicle component detection for obstacle awareness in AEB systems, for an autonomous vehicle to safely navigate in traffic, learning how to detect surround vehicle lights is a crucial component toward understanding meaningful information (like turn signals and brake indications) that assist in predicting the future actions of other traffic agents (such as lane or speed changes). From these detected tail lights, we suggest that future research should seek to derive these temporal signals (such as turn and brake indications), ultimately serving the task of safe path planning. 


{\small
\bibliographystyle{elsarticle-num.bst}
\bibliography{refs}
}

\end{document}